\documentclass[letterpaper]{article} 

\usepackage{aaai23}  
\usepackage{times}  
\usepackage{helvet}  
\usepackage{courier}  
\usepackage[hyphens]{url}  
\usepackage{graphicx} 
\usepackage{natbib}  
\usepackage{caption} 
\usepackage{algorithm}
\usepackage{algorithmic}
\usepackage{booktabs}
\usepackage{multirow}
\usepackage{graphicx}
\usepackage{amssymb}
\usepackage{amsmath,amssymb,amsfonts}
\usepackage{newfloat}
\usepackage{listings}
\DeclareCaptionStyle{ruled}{labelfont=normalfont,labelsep=colon,strut=off} 
\floatstyle{ruled}
\newfloat{listing}{tb}{lst}{}
\floatname{listing}{Listing}
\usepackage{threeparttable}
\usepackage{booktabs}
\usepackage{multirow}
\usepackage{colortbl}
\usepackage[table,xcdraw]{xcolor}
\usepackage{multirow}

\title{TripleSurv: Triplet Time-adaptive Coordinate Loss for Survival Analysis}

\author{
    Liwen Zhang,\textsuperscript{\rm 1}\equalcontrib
    Lianzhen Zhong,\textsuperscript{\rm 1,\rm 2}\equalcontrib
    Fan Yang,\textsuperscript{\rm 1}
    Di Dong,\textsuperscript{\rm 1}
    Hui Hui,\textsuperscript{\rm 1}
    Jie Tian\textsuperscript{\rm 1,\rm 3}\thanks{Corrsponding author.}\\
}
\affiliations{
    \textsuperscript{\rm 1}CAS Key Laboratory
of Molecular Imaging, Institute of Automation, Beijing, China\\
\textsuperscript{\rm 2} China Mobile (Hangzhou) Information Technology Co. LTD., Hangzhou, China
    \textsuperscript{\rm 3} School of Engineering Medicine \& School of Biological Science and Medical Engineering, Beihang University\\
    
    zhangliwen2018, zhonglianzhen2018, yangfan2022, HuiHui@ia.ac.cn, jie.tian@ia.ac.cn

}

\usepackage{bibentry}

\begin{document}

\maketitle

\begin{abstract}
A core challenge in survival analysis is to model the distribution of censored time-to-event data, where the event of interest may be a death, failure, or occurrence of a specific event. Previous studies have showed that ranking and maximum likelihood estimation (MLE) loss functions are widely-used for survival analysis. However, ranking loss only focus on the ranking of survival time and does not consider potential effect of samples’ exact survival time values. Furthermore, the MLE is unbounded and easily subject to outliers (e.g., censored data), which may cause poor performance of modeling. To handle the complexities of learning process and exploit valuable survival time values, we propose a time-adaptive coordinate loss function, TripleSurv, to achieve adaptive adjustments by introducing the differences in the survival time between sample pairs into the ranking, which can encourage the model to quantitatively rank relative risk of pairs, ultimately enhancing the accuracy of predictions. Most importantly, the TripleSurv is proficient in quantifying the relative risk between samples by ranking ordering of pairs, and consider the time interval as a trade-off to calibrate the robustness of model over sample distribution. Our TripleSurv is evaluated on three real-world survival datasets and a public synthetic dataset. The results show that our method outperforms the state-of-the-art methods and exhibits good model performance and robustness on modeling various sophisticated data distributions with different censor rates. Our code will be available upon acceptance.
\end{abstract}

\section{1. Introduction}

	Survival analysis is a set of techniques to analyze data related to the duration of time until an event of interest occurs \cite{RN991}. This approach is applied in several fields including medicine, engineering, economics, and sociology. The purpose of survival analysis is to assess the impact of certain variables on survival time \cite{RN305}. As an example, a survival analysis can be used to investigate the relationship between clinical factors (e.g., age, gender, and race) and heart attack risk. Additionally, survival analysis is frequently utilized in the medical field to identify which factors have the greatest impact on disease recurrence \cite{RN991}.
	
Standard statistical and machine learning methods are widely used for survival analysis. Cox Proportional Hazards (CPH) \cite{RN644} is one of prevalent models, which calculates the effects of variables on the risk of an event happening. The CPH model is based on the assumption that patients' risk of death is a linear combination of their variables. However, this assumption is too strong to reflect the real-world nonlinear relationships between survival time and variables. Recently, researchers have attempted to improve the performance of survival analysis models by incorporating deep learning techniques to augment the conventional Cox model \cite{RN645}. Previous study introduce a deep generative model within the framework of parametric censored regression \cite{RN1021}. However, these deep survival analysis models still possess strong assumptions of the CPH model.

To address the limitations of previous deep survival analysis models, discrete-time survival analysis based on Maximum Likelihood Estimation (MLE) draws much attention \cite{RN632deephit}. This approach segments the observation period into multiple time intervals, and then predicts survival time by determining if the event of interest has occurred at certain interval. For instance, a deep neural network (DNN) was proposed in \cite{RN632deephit} that combines ranking loss and likelihood loss to predict the probability density values for discrete-time survival analysis. Additionally, techniques such as multi-task learning algorithm \cite{RN1022} and recurrent neural networks (RNN) \cite{RN651}, the studies have been used to capture the relationships between adjacent time intervals. 

The loss function is a crucial component of survival model learning. Some studies have optimized survival models using ranking loss by predicting the order of survival times in pairs of samples \cite{RN632deephit}. However, the original ranking loss function only focuses on the order of predicted values rather than the specific values themselves, and disregarding the quantitative differences of survival time for individuals. Besides, some researchers currently focus on the calibration performance of survival model \cite{RN902}. Calibration refers to that predictions are consistent with observations, a well-calibrated survival model is one where the predicted probabilities over events within any time interval are consistent with the observed frequencies of their occurrence \cite{RN993}. Although these methods do not make any strong assumptions about the underlying distribution of survival time or survival function, they still have some limitation: 1) Not consider potential effect of samples’ exact survival time values; 2) improving the model performance from a single aspect, without revealing the relationship between accuracy and robustness of models. To address these issues, we propose a novel loss function, TripleSurv, to further optimize the modeling process from different perspectives. Main contributions of our work are summarized as follows: 

\begin{itemize}
\item We propose a time-adaptive pairwise loss function to exploit valuable survival information, and achieve adaptive adjustments by introducing the differences in the survival time between sample pairs into the ranking, which can encourage the model to quantitatively rank relative risk of pairs, ultimately enhancing the accuracy of predictions.
\item We propose a coordinate loss function of TripleSurv to optimize the modeling process. The TripleSurv strikes a balance to facilitate the model to takes into account the data distribution, ranking, and calibration.
\item Our TripleSurv is evaluated on four public datasets. The results demonstrate that our method outperforms the state-of-the-art and existing methods and achieves good performance and robustness on modeling various sophisticated data distributions with the highest geometrical and clinical metrics. Most importantly, our method also performs well on datasets with large censoring rates.

\end{itemize}

\section{2. Related work}
We review three streams of related work for survival loss function in terms of the technical components of this work: 1) Likelihood estimation; 2) Ranking; 3) Calibration. A brief summary can be found in Appendix A.

\subsection{2.1. Methods Based on Likelihood}
Likelihood estimation function is a commonly used to optimize survival analysis models. One of representative Likelihood estimation functions is Maximum Likelihood Estimation (MLE). Though MLE corresponds to a proper scoring rule for modeling distribution, it is sensitive to outliers \cite{RN909}. This sensitivity may result in poor generalization of model. Continuous Ranked Probability Score (CRPS) is a great substitution of MLE which has been widely used in meteorology \cite{RN5}. The CRPS gives more calibrated forecasts compared with MLE. More importantly, CRPS could imporve the sharpness of probabilistic forecast which is more practical for survival models \cite{RN1021,RN6,RN7}. Avati et al. introduced a Survival-CRPS (S-CRPS) for survival analysis which extended CRPS to handle right and interval-censored data. Nevertheless, although S-CRPS shows good robustness, it does not gives a straightforward way to balance model performance between the discrimination and robustness.
\subsection{2.2. Methods Based on Ranking}
Ranking for survival analysis is a statistical method used to analyze the time-to-event data, where the events of interest are ranked or ordered. The Concordance Index (C-index) is a widely used metric for ranking \cite{RN408}. The C-index focuses on ranking problem which calculates a ratio of corrected ordered pairs among all possible comparable pairs. However, it can’t be directly used as an objective function during training for it is invariant to any monotone transformation of the survival times \cite{RN8}. To overcome this problem, a number of related works have emerged considering ranking problem by introducing ranking loss in-train to improve the ranking ability of survival models. Additionally, Cox’s partial likelihood function \cite{RN9} is commonly used as the objective function in Cox proportional hazard model \cite{RN645, RN10}, which describes the risk of an event occurring for an individual at a specific time point, given certain covariates. Raykar et al. proved that maximizing this likelihood also ends up approximately maximizing the C-index. Recently, many works attempt to improve C-index by combining ranking loss function in-training \cite{RN632deephit,RN12,RN991}. The study \cite{RN632deephit} proposed a deep learning method for modeling the event probability without assumptions of the probability distribution by combining MLE with a ranking loss. In RankDeepSurv \cite{RN991}, the authors combine a selective ranking loss with MSE. However, these work mainly focus on ordering relationship between comparable pairs but ignore the specific numerical differences for survival time.
\subsection{2.3. Methods Based on Calibration}
	Calibration refers to that predictions are consistent with observations, a well-calibrated survival model is one where the predicted probabilities over events within any time interval is consistent to the observed frequencies of their occurrence \cite{RN993}. Survival models with poor calibration can cause poor generalization for predicting the distribution if real-world survival data \cite{RN16,RN17}. Recently, many studies deal with calibration problems in-training for survival model. Avati et al. \cite{RN993} replaced the common used partial likelihood loss with Survival-CRPS, which could implicitly balance between prediction and calibartion. Kamran proposed rank probability score(RPS) which is a discrete approximation based on CRPS as well. Concurrently, Goldstein et al. proposed an explicit differentiable calibration loss of X-CAL \cite{RN902} for boosting model robustness in-training. However, X-CAL doesn't disclose the relationship between the discrimination and robustness.

\section{3. Methods}
In this section, we mainly introduce survival data, data preprocessing, methodology of our proposed method, and evaluation metrics.
\subsection{3.1. Survival Data}
Survival data consist of three pieces of information  $(\vec X, t,\delta,)$ for each sample: 1) The vector $\vec X $ denotes available covariates; 2) observed survival time $t$ elapsed between enrollment time and the time of the failure or the censoring, whichever occurred first; 3) a label indicating the status of event $\delta$ (e.g. recurrence or death). One peculiar feature for survival data is known as censoring. A censored sample signifies that a patient did not experience the failure during the observed time interval.

\subsection{3.2. Data Preprocessing}
\begin{figure}[htbp]
    \centering
    \includegraphics[width=0.47\textwidth]{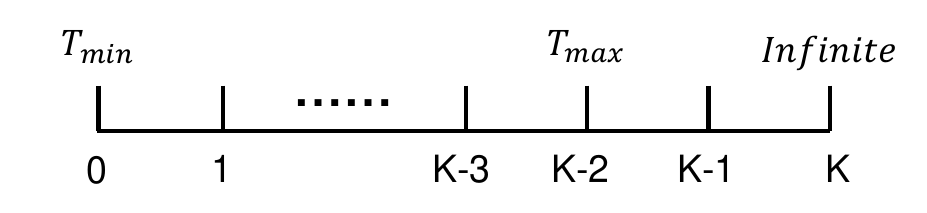}
    \caption{Diagram of the interval partition of study time.}
    \label{fig1-time}
    \end{figure}

To standardize the survival time, we normalize it to a range of 0 to 1. We also define $T_{max}=max\{t_{i}|\delta_{i}=1\}$, and $T_{min}=min\{t_{i}|\delta_{i}=1\}$. Since the survival times of censored samples may be greater than $T_{max}$, and the last interval needs to be left to correspond to infinity, $T_{max}$ corresponds to K-2,$T_{min}$corresponds to 0. The diagram is shown as Figure \ref{fig1-time}.

To leave a certain interval, in fact, we make $T_{max}$ correspond to K-2.1, $T_{min}$ corresponds to 0.1, so the time interval of each interval $\Delta T=\frac{T_{max}-T_{min}}{K-2.2}$, and then the study time $t$ is normalized according to follow formulas:
\begin{equation}
crop(x,a,b)=
\left\{
             \begin{array}{lr}
             a,x<a &  \\
             b,x>b\\
             x,other &  
             \end{array},
\right.
\end{equation}
\begin{equation}
\left\{
             \begin{array}{lr}
             T_{min}^{'}=T_{min}-0.1*\Delta T  \\
             T_{max}^{1}=T_{min}^{'}+(K-1) * \Delta T & \\
             T_{max}^{2}=T_{min}^{'}+K*\Delta T & \\
             t^{'}=crop(t,T_{min}^{'},T_{max}^{1}) & \\
             \tilde{t}=(t^{'}-T^{'}_{min})/(T_{max}^{2}-T_{min}^{'})& 
             \end{array},
\right.
\end{equation}
where the width corresponding to each time interval after normalization is $\frac{1}{K-1}$. The k-th interval represents the time interval $[\frac{k-1}{K},\frac{k}{K})$, corresponding time $t^{bin}_{k}=\frac{2k-1}{2K}$. Unless otherwise specified, the study time $t$ refers to the time after normalization in the following context.

\subsection{3.3. Proposed TAPR-loss and TripleSurv}
We introduce our proposed loss function in two parts: 1) Theoretical description of our proposed time-adaptive pairwise rank loss (TAPR-loss); 2) Theoretical description of our proposed integrating TripleSurv equipped with TAPR-loss.

\subsubsection{Proposed TAPR-loss}
Our proposed time-adaptive pairwise ranking loss (TAPR-loss) is inspired by ranking loss \cite{RN632deephit}. The ranking loss function optimizes directly the pairwise orders: the longer-lived one among comparable pair have a lower risk than the other:
\begin{equation}
l_{rank}=\frac{1}{|A^{1}|}\sum_{i \in A^{1}}\sum_{j \in R(i)}\Pi(risk_{i}-risk_{j}),
\end{equation}
where $A^{1}=\{i|\delta _{i}=1\}$, $R(i)$ is called the risk set and is defined as $R(i)=\{ j|t_{j}>t_{i} \}$, $risk_{i}$ signifies the estimated risk of the sample $i$, $\Pi(\cdot)$ is the indicator function, and $|\cdot|$ is a counting function for a set.
Since the indicator function is non-differentiable, we generally use a function $\eta(\cdot)$ to approximate the indicator function. Lee et al. \cite{RN632deephit} used $\eta(risk_{i},risk_{j})=exp(\sigma*(risk_{i}-risk_{j}))$ and rewrote the rank loss as follows:
\begin{equation}
\begin{aligned}
l_{rank}=\frac{1}{|A^{1}|}\sum_{i \in A^{1}}\sum_{j \in R(i)}exp(\sigma*(F(t_{i}|\vec X_{i})-F(t_{i}|\vec X_{j})))
\end{aligned},
\label{eqrank}
\end{equation}
where $F(t_{i}|\vec X_{i})$ signifies the probability of failure occurring at time $t_{i}$ and the given covariates $\vec X_{i}$. $\sigma$ is a scalar hyperparameter.
The rank loss only considers the relative ranking of the survival time but ignore the important quantitative difference \cite{RN632deephit,RN360}, resulting in countless optimal solutions for the optimization problem, even invalid forecasts. To mitigate the issues, we extend the idea of concordance and assume that the risk difference between the comparable pair is proportional to the difference of failure times between them.  According to the assumption, we add the difference of failure times into the rank loss, forming the TAPR-loss:
\begin{displaymath}
l_{TAPR-loss}=
\end{displaymath}
\begin{equation}
\frac{1}{|A^{1}|}\sum_{i \in A^{1}}\sum_{j \in R(i)}exp(\sigma*[(risk_{i}-risk_{k})-\rho*(t_{j}-t{i})]),
\end{equation}
where we refine $risk = 1-mean\enspace dead\enspace time = 1- \sum_{k-1}^{K}p_{k}*t^{bin}_{k}$, $\sigma$ and $\rho$ are scalar hyperparameters, $\sigma \in (0,1]$. Considering $t^{bin}_{k}\in [\frac{1}{2K},\frac{2K-1}{2K}]$, we can infer $\frac{1}{2K}=1-\frac{2K-1}{2K}\sum_{k=1}^{K}p_{k}<risk \leq 1-\frac{1}{2K}\sum_{k=1}^{K}p_{k}=\frac{2K-1}{2K}$, and $risk \in [\frac{1}{2K},\frac{2K-1}{2K}]$.

\subsubsection{TripleSurv: strike a trade-off time-adaptive coordinate loss}
To improve the performance and robustness of model and exploit valuable survial time, we propose the TripleSurv to optimize the modeling process in multiple scale (single-sample, pairs, and population level) by integrating the likelihood loss, TAPR-loss, and calibration loss:
\begin{equation}
\begin{aligned}
l_{TripleSurv }= & -\alpha * l_{likelihood } \\
& -\beta * l_{TAPR-loss }+\gamma * l_{calibration }
\end{aligned},
\end{equation}

where $\alpha$, $\beta$ and $\gamma$ are scalar hyperparameters, which are suggested to set for ensuring that the values of these three items are at the same level of magnitude.
\newline
\newline
\subsubsection{Likelihood loss ($l_{likelihood}$)}

Theoretically, we need to estimate probability density function $f(t|\vec X)$ and the likelihood can be written as follows:
\begin{equation}
l_{likelihood}=
\left\{
             \begin{array}{lr}
            f(t|\vec X), \delta =1 &  \\
            S(t|\vec X), \delta =0 &  
             \end{array},
\right.
\end{equation}
where $S(t|\vec X)$ is survival function. $\delta =1$ represent event status is observed while $\delta =0$ represent event status is not observed. In this study, we use the discrete probability mass function $P(t|\vec X)=[p_{1},p_{2},...,p_{k}]$ in $k$ disjoint time intervals, which is often predicted in academic research and clinical practical, and its definition and the likelihood can be written as follows:
\begin{equation}
l_{likelihood}=
\left\{
             \begin{array}{lr}
            p_{k}, \delta =1 &  \\
            1-\sum_{i=1}^{k}, \delta =0 &  
             \end{array},
\right.
\end{equation}
where $p_{k},k=1,2,...,K$, signifies the probability of the failure occurring in a specific time interval $[a_{k},b_{k})$, $k$ denotes the index of time interval that study time $t$ falls.

\subsubsection{Calibration loss ($l_{calibration}$ )}
We combine the calibration loss with other objectives for optimization during training to strike a desired balance between discrimination and robustness.
\begin{equation}
\left\{
         \begin{array}{lr}
        l_{calibration}=\frac{1}{G}\sum^{G}_{g=1}(pred_{g}-obse_{g})^{2} &  \\
        pred_{g}=(\sum_{i}\sum_{t^{bin}_{k} \in I_{g}} p^{i}_{k} )/ (\sum_{i}\sum_{t^{bin}_{k}>a_{g}} p^{i}_{k}) &  \\
         obse_{g}=| \{ i|a_{g} \leq t_{i} < b_{g} , e_{i}=1\}| /| \{ i|a_{g} \leq t_{i} \} |                    &
         \end{array}.
\right.
\end{equation}

We compute the squared errors between the observed and predicted failure incidence within $G$ time intervals. Obviously, the optimization of calibration constrains the distribution of the model prediction in population level, which can play an important role of regularization in training process.

\subsection{3.4. Model Description}
We exploit Categorical (Cat) and Multi-Task Logistic Regression (MTLR) methods \cite{RN902} equipped with our proposed loss to model the distribution of failure occurring over discrete times. The Cat method is parameterized by using a deep neural network function of with $K$ bins as outputs, which can approximate the continuous survival distribution as $K \to \infty$. MTLR method is similar to the Cat method except that it allows the model to produce $K-1$ outputs. Assume a survival model with MTLR method outputs a vector $\varphi =[\varphi_{1},\varphi_{2},...,\varphi_{K-1}]$, the estimation of the probability mass function for bin $k<K$ is:
\begin{equation}
p_{k}=\frac{exp(\sum_{j=k}^{K-1}\varphi_{j})}{1+\sum_{i=1}^{K-1}exp(\sum_{j=i}^{K-1}\varphi_{j})},
\end{equation}
and the estimation of the probability mass function for bin $K$ is:
\begin{equation}
p_{k}=\frac{1}{1+\sum_{i=1}^{K-1}exp(\sum_{j=i}^{K-1}\varphi_{j})}.
\end{equation}

\begin{figure}[htbp]
    \centering
    \includegraphics[width=0.47\textwidth]{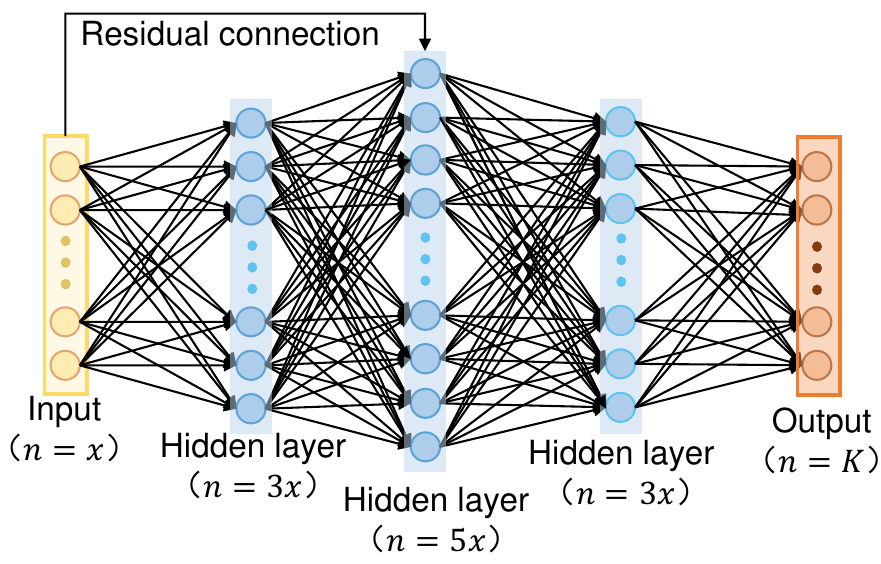}
    \caption{The fully-connected neural network for one-dimensional data.}
    \label{fig2arch}
    \end{figure}

As for one-dimensional data, we use a four layer fully-connected residual neural network (Figure \ref{fig2arch}) as the architecture of the survival models, which is similar to the proposed model by Lee et al \cite{RN632deephit}. For the public synthetic dataset, we use shallow ResNet \cite{RNhe2016deep} as the architecture. The Batch Normalization is used in the architectures.
\subsection{3.5. Evaluation Metrics}
We use popular metrics of Concordance index (C-index), Brier Score, and Time-dependent area under the ROC (TDAUC) for evaluation of model performance from different aspects. Moreover, in our study, risk stratification for real-world data is also conducted to assess the benefit of the survival models for selection of decision-making. We introduce the definition of these metrics for easy to reproduce in Appendix B.

\subsubsection{Concordance Index (C-index)}
The C-index is a widely-used metric for survival models, which severs as a representative indicator for rank relationship between predicted risk scores $\hat r$ and observed time points $t$. A C-index of 1.0 indicates perfect discrimination, while a C-index of 0.5 represents no discrimination ability.

\subsubsection{Integrated Brier Score}
The Integrated Brier Score (IBS) is an extension of the time-dependent Brier score for survival model with cencored data. Time-dependent Brier score is a tailored metric is extended by Graf et al. \cite{RN1038} and widely-used to measures the capability of calibration. The metric uses inverse probability of censoring weights, which requires estimating the censoring survival function, denoted as $\hat G(t)$ over time points $t$. The time-dependent Brier score is defined as:
\begin{equation}
\left\{\begin{array}{c}
I_i^{\text {sum }}(t^*)=I\left(t_i \leq t^* \& \delta_i=1\right) \frac{\left(0-\hat{S}\left(t^*\right)\right)^2}{\hat{G}\left(t_i\right)}\\
+I\left(t_i>t^*\right) \frac{\left(1-\hat{S}\left(t^*\right)\right)^2}{\hat{G}\left(t^*\right)} \\
B S\left(t^*\right)=\frac{1}{n} \sum_{i=1}^n I_i^{\text {sum }}(t^*)
\end{array},\right.
\end{equation}
where $I(\cdot)$ represents observed event status, $n$ is number of samples, and $\hat S(t)$ is the observed rate of event-free samples at $t^*$. IBS can be defined given an average IBS across the time in intervals of $T^*$:

\begin{equation}
\operatorname{IBS}(T^*)=\frac{1}{T^*} \int_0^{T^*} BS(t^*) dt^*.
\end{equation}

\subsubsection{Time-dependent area under the ROC (TDAUC)}
The metirc of TDAUC is a performance evaluation metric for binary classifiers that takes into account the classifier's performance changes over time \cite{RN1036}. At a given time point $t$ and a cutoff $c$, We define them as:

\begin{displaymath}
TDAUC(t,X)=
\end{displaymath}
\begin{equation}
Aera\{sensitivity(c,t),1-specificity(c,t)\}.
\end{equation}
We use the mean of TDAUC (mTDAUC) over the time to determine how well estimated risk scores can distinguish diseased patients from healthy patients:
\begin{equation}
mTDAUC(X)=\frac{1}{n_{t}}\sum TDAUC(t_{i},X),
\end{equation}
where $n_{t}$ denote the number of time points.

\subsubsection{Hazard Ratio}
The hazard ratio (HR) is a measure of the relative risk of an event occurring in one group compared to another group over time. To evaluate the effectiveness of different decision-making, decision implementation mainly is based on risk stratification with HR, where samples are divided into low-risk and high-risk groups by a risk factor $r$ and are recommended a favorable regimen. The cutoff of the risk factor is determined by the partition with the maximal log-rank test statistic in the training set.

\section{4. Experiments and Results}
\subsection{4.1. Competing Methods}

\subsubsection{a. Lasso-Cox}
The Cox proportional hazards (CPH) model is the most frequently used survival model \cite{RN991}.  Generally, to avoid wrong estimation caused by redundant input features, the Least Absolute Shrinkage and Selection Operator (LASSO) is used to perform feature selection before building CPH model. They are often used together.
\subsubsection{b. RSF}
The Random Survival Forests (RSF) is a popular nonlinear survival model that is based on the tree method and produces an ensemble estimate for the cumulative hazard function.
\subsubsection{c. DeepSurv}
DeepSurv model \cite{RN645} is a CPH model based on full connect neural network that optimizes the partial likelihood loss and outputs directly for the survival risk prediction.
\subsubsection{d. DeepRank}
We formulate the DeepRank model as the comparison model that optimizes the rank loss \cite{RN632deephit}.
\subsubsection{e. CRSP}
Survival-CRPS (CRPS) \cite{RN993} is sharpness subject to calibration but it optimizes neither the calibration loss nor the traditional likelihood loss.
\subsubsection{f. X-cal}
X-cal model \cite{RN902} combines the likelihood loss and explicit calibration for estimating the distribution of survival function.
\subsubsection{g. DeepHit}
The DeepHit model \cite{RN632deephit}, introduced by Lee et al. in 2018, is a deep neural network that predicts the probability $p(z)$ of an event occurring over the entire time space given an input $x$. This method exhibits state-of-the-art performance for modelling distribution of time-to-event data.
\subsubsection{h. Cat integrating different loss functions}
Cat is an abbreviation of categorical method \cite{RN902}. This method regards survival analysis as a classification task, which discretizes the survival time of patients into several time interval bins, and then predicts the probability of each or falling in its corresponding interval bins. In our study, we compared different Cat methods integrating CRSP (Cat-crps), X-cal (Cat-xcal), DeepHit (Cat-hit), and our method (Cat-ours).
\subsubsection{i. MTLR integrating different loss functions}
MTLR \cite{RN1022} method is different from the Cat method considering some relationships between the probability of the time interval bins. We also compared our method (MTLR-ours) integrating our proposed loss function with MTLR-crps, MTLR-xcal, and MTLR-hit.

\subsection{4.2. Implementation Details}
We use SGD optimizer and the “cosine annealing” learning schedule to update training weights. The initial learning rate is set as 1e-3 or 1e-2 where appropriate. The rate of dropout is set as 0.2 unless otherwise specified. The weight of each component in TripleSurv loss, $\alpha$, $\beta$, and $\gamma$, are set to ensure the same level of magnitude for their values in the training datasets. The $\sigma$, $\rho$ are determined according to the model performance in the validation dataset. In each experiment setting, the final model used for model evaluation was determined using its performance in terms of the C-index in the validation. The default ratio of training, validation, and test sets is approximately 3:1:1. For the small sample size of METABRIC (n=1981), we employ five-fold cross-validation for performance evaluation of different methods. More details for easy reproduction in Appendix C.

\subsection{4.3. Datasets and Tasks}
\subsubsection{SUPPORT}
The Study to Understand Prognoses Preferences Outcomes and Risks of Treatment (SUPPORT) is a comprehensive study that assessed the survival time of critically ill adults who were hospitalized \cite{RN991}. The SUPPORT comprises 9105 patients and encompasses 14 different features. In the dataset, the censor rate is 31.9\%. Totally 68.10\% of the dataset was observed data.

\subsubsection{BIDDING}
BIDDING real-time bidding dataset that contains auction request information, bid price, and the status whether bidders win the auction. Researchers treat the bid price as the time and whether winning of the auction as the event status for survival analysis, and treat winning probability estimation of a single auction as a task \cite{RN651}.

\subsubsection{METABRIC}
We evaluate different methods for the prediction of overall survival of patients with breast cancer from the Molecular Taxonomy of Breast Cancer International Consortium (METABRIC) dataset \cite{RN632deephit}. Totally 1980 patients are avaliable in the dataset, which contains gene expression profiles and clinical features \cite{RN991}. Among all patients, 888 (44.8\%) were followed until death, while the remaining 1093 (55.2\%) were right-censored.
\subsubsection{MINIST}
Moreover, a public synthetic dataset \cite{RN902} based on the MNIST is considered for evaluating our proposed method. The synthetic survival times are conditional on the MNIST classes.The settings for MNIST are the same as \cite{RN902}.  

More details for statistics of all datasets are summarized in Appendix D.

\subsection{4.4. Results and Analysis}
Four experiments are conducted and show that our proposed TripleSurv performs well on datasets with different censoring rates. Compared to existing loss functions of MLE \cite{RN651}, rank loss \cite{RN632deephit}, and calibration \cite{RN902}, our TripleSurv achieves the best ranking accuracy and robustness on four datasets. More details for prognostic risk evaluation can be found in Appendix E.

\subsubsection{Experiment 1: Overall survival prediction in SUPPORT}

\begin{table}[ht]
	\centering
	\resizebox{\linewidth}{!}{
	\begin{tabular}[c]{ccccc}
		\toprule
		{Method} & {C-index($\uparrow$)} & {mTDAUC($\uparrow$)} & {IBS($\downarrow$)} & {HR($\uparrow$)} \\
		\midrule
		Random(ref) & 0.500 & 0.500 &0.2520& 1.00 \\
		Lasso-cox&  0.725&	0.754 &	0.1846 & 2.92 \\
		RSF&	    0.713& 	0.762&	0.1806& 	2.99 \\
		DeepSurv&	0.721& 	0.757&	0.1867 &	3.00 \\
		DeepRank&	0.723&	0.742&	0.1906 &	2.77 \\
		Cat-crps&	0.675& 	0.684&	0.1999& 	2.19 \\
		Cat-xcal&	0.704& 	0.728&	0.1908& 	2.55 \\
		Cat-hit&	0.712& 	0.757&	0.1843& 	3.01 \\
		Cat-ours&	0.726& 	0.762&	\textbf{0.1803}& 	3.04 \\
		MTLR-crps&	0.701& 	0.724&	0.1918& 	2.53 \\
		MTLR-xcal&	0.701& 	0.744&	0.1892& 	2.69 \\
		MTLR-hit&	0.711& 	0.761&	0.1840& 	\textbf{3.07} \\
		MTLR-ours&	\textbf{0.727}& 	\textbf{0.765}&	0.1804& 	3.04 \\
		\bottomrule
	\end{tabular}}
	\caption {Performance comparison in SUPPORT. C-index:Concordance Index; mTDAUC: mean Time-dependent area under the ROC; IBS: Integrated Brier Score; HR: hazard ratio (clinical metric for risk evaluation).}
	\label{tab:chap:table3Supp}
\end{table}

\begin{figure}[!ht]
    \centering
    \includegraphics[width=0.47\textwidth]{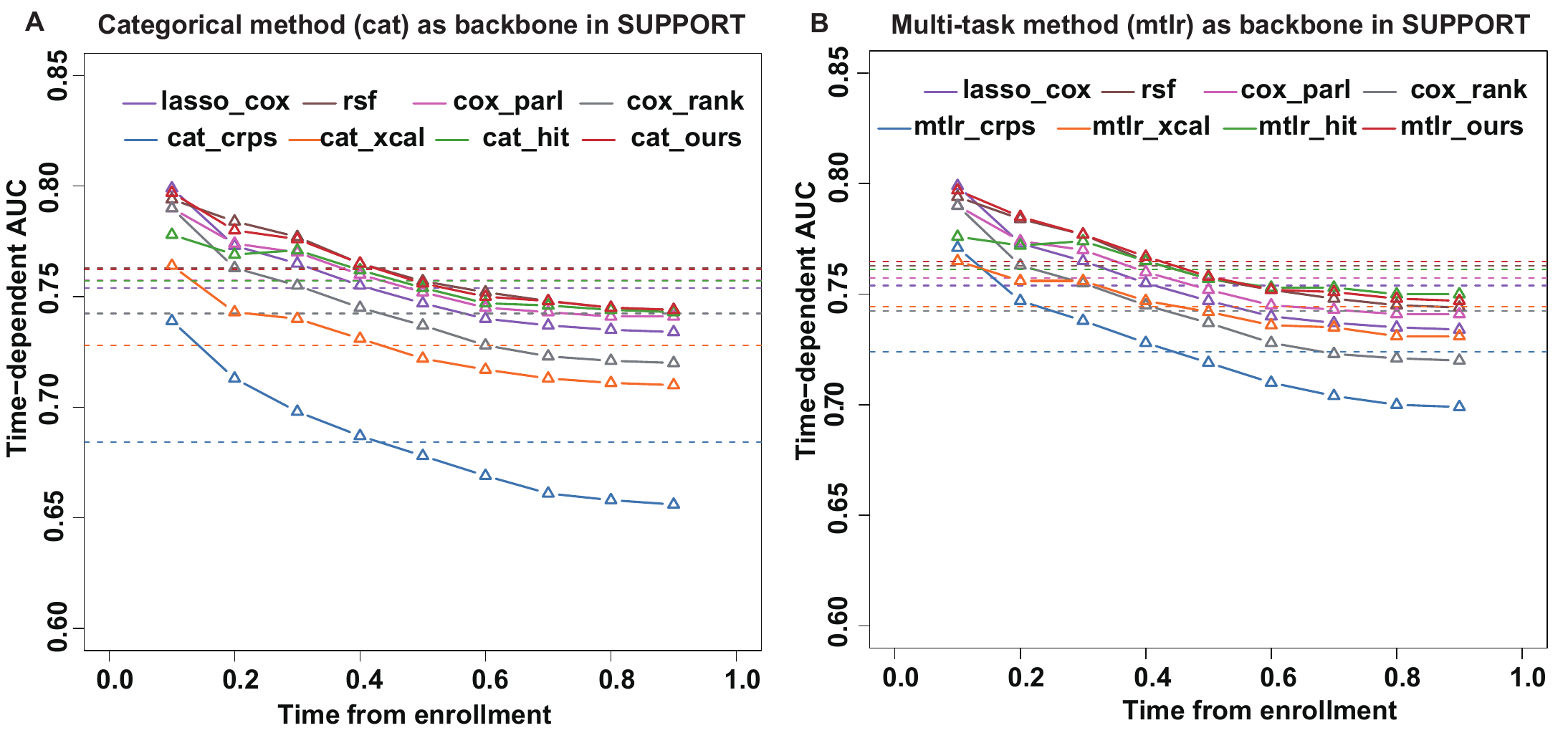}
    \caption{Performance comparison using the TDAUC in SUPPORT.}
    \label{fig:chap:fig3-supportAUC}
    \end{figure}

As shown in Table \ref{tab:chap:table3Supp}, compared to existing deep survival models, the Cat-ours achieved the highest C-index, mTDAUC (the evaluation of TDAUC at each time point is shown in Figure \ref{fig:chap:fig3-supportAUC}), and the lowest IBS. The same results were also observed in MTLR-ours. The results indicate that our proposed TripleSurv has excellent discriminative ability and can achieve a good balance between discriminative ability and robustness for survival models. 

\subsubsection{Experiment 2: Winning prediction at auction in BIDDING}

\begin{table}[htbp]
	\centering
	\resizebox{\linewidth}{!}{
	\begin{tabular}[c]{ccccc}
		\toprule
		{Method} & {C-index($\uparrow$)} & {mTDAUC($\uparrow$)} & {IBS($\downarrow$)} & {HR($\uparrow$)} \\
		\midrule
		Random (ref)&	0.500 &	0.500 &	0.2926 &	1.00  \\
		Lasso-cox&	0.699&	0.716&	0.2234&	3.07 \\
		RSF&	0.766&	0.791&	0.1945&	3.66 \\
		DeepSurv&	0.762&	0.778&	0.1990&	3.65 \\
		DeepRank&	0.760&	0.777&	0.2091&	3.69 \\
		Cat-crps&	0.680&	0.695&	0.2900&	2.41 \\
		Cat-xcal&	0.723&	0.742&	0.2009&	2.92 \\
		Cat-hit&	0.761&	0.783&	0.2061&	3.52 \\
		Cat-ours&	0.782&	0.800&	\textbf{0.1922}&\textbf{3.95} \\
		MTLR-crps&	0.751&	0.767&	0.2392&	3.14 \\
		MTLR-xcal&	0.727&	0.750&	0.1997&	3.02 \\
		MTLR-hit&	0.773&	0.791&	0.2078&	3.59 \\
		MTLR-ours&	\textbf{0.785}&	\textbf{0.801}&	0.1935&	3.76 \\
		\bottomrule
	\end{tabular}}
	\caption { Performance comparison in BIDDING}
	\label{tab:chap:tablebidding}
\end{table}

\begin{figure}[!ht]
    \centering
    \includegraphics[width=0.47\textwidth]{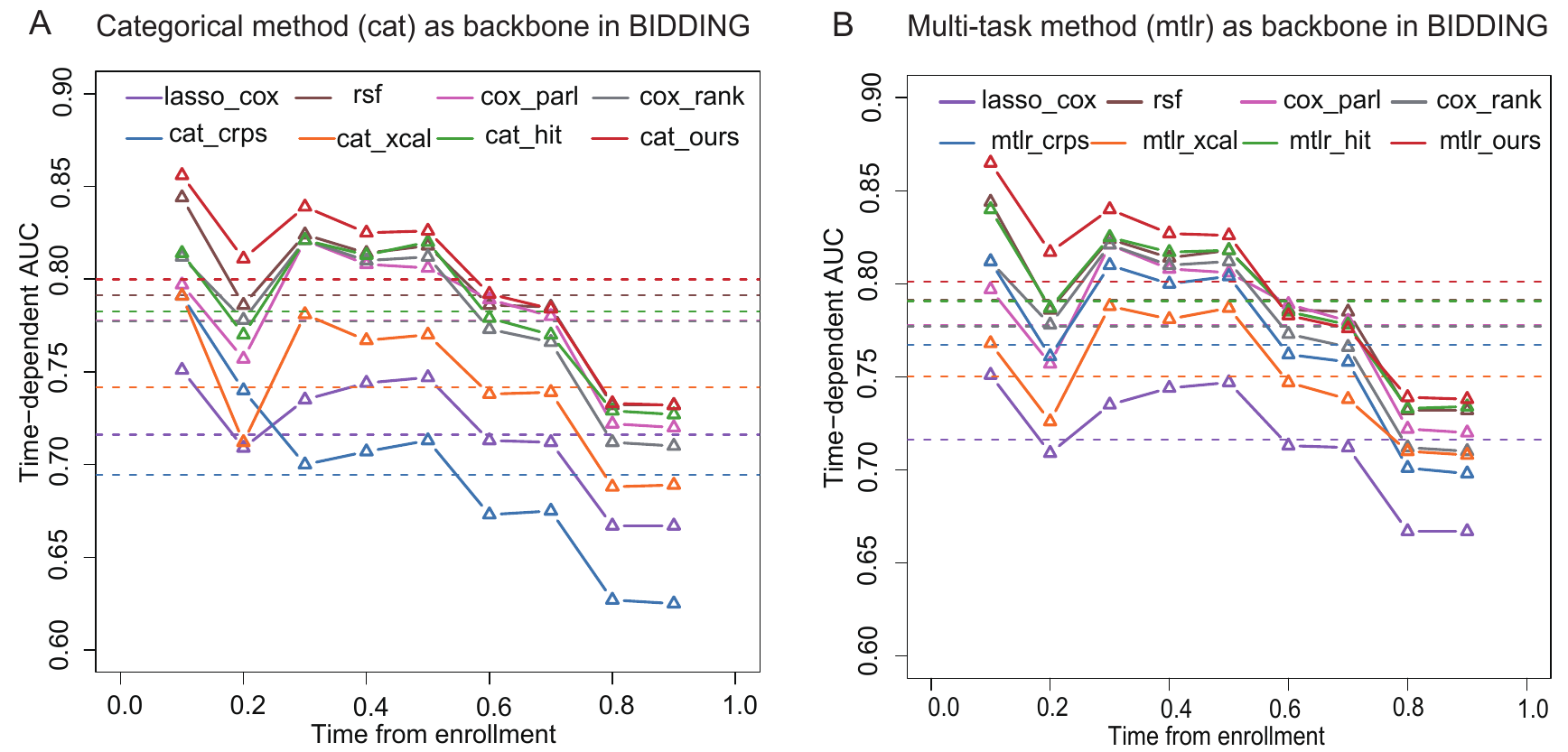}
    \caption{Performance comparison using the TDAUC in BIDDING.}
    \label{fig5:bidAUC}
    \end{figure}

Our models (Cat-ours and MTLR-ours) have the best discriminative and calibration capability (Table \ref{tab:chap:tablebidding} and Figure \ref{fig5:bidAUC}). For survival models based on CRPS loss, the results show a significant difference between Category and MTLR methods. The performance of nonlinear survival models is generally better than linear survival models (Lasso-cox). 

\subsubsection{Experiment 3: Overall survival prediction in METABRIC}
\begin{figure}[ht]
    \centering
    \includegraphics[width=0.47\textwidth]{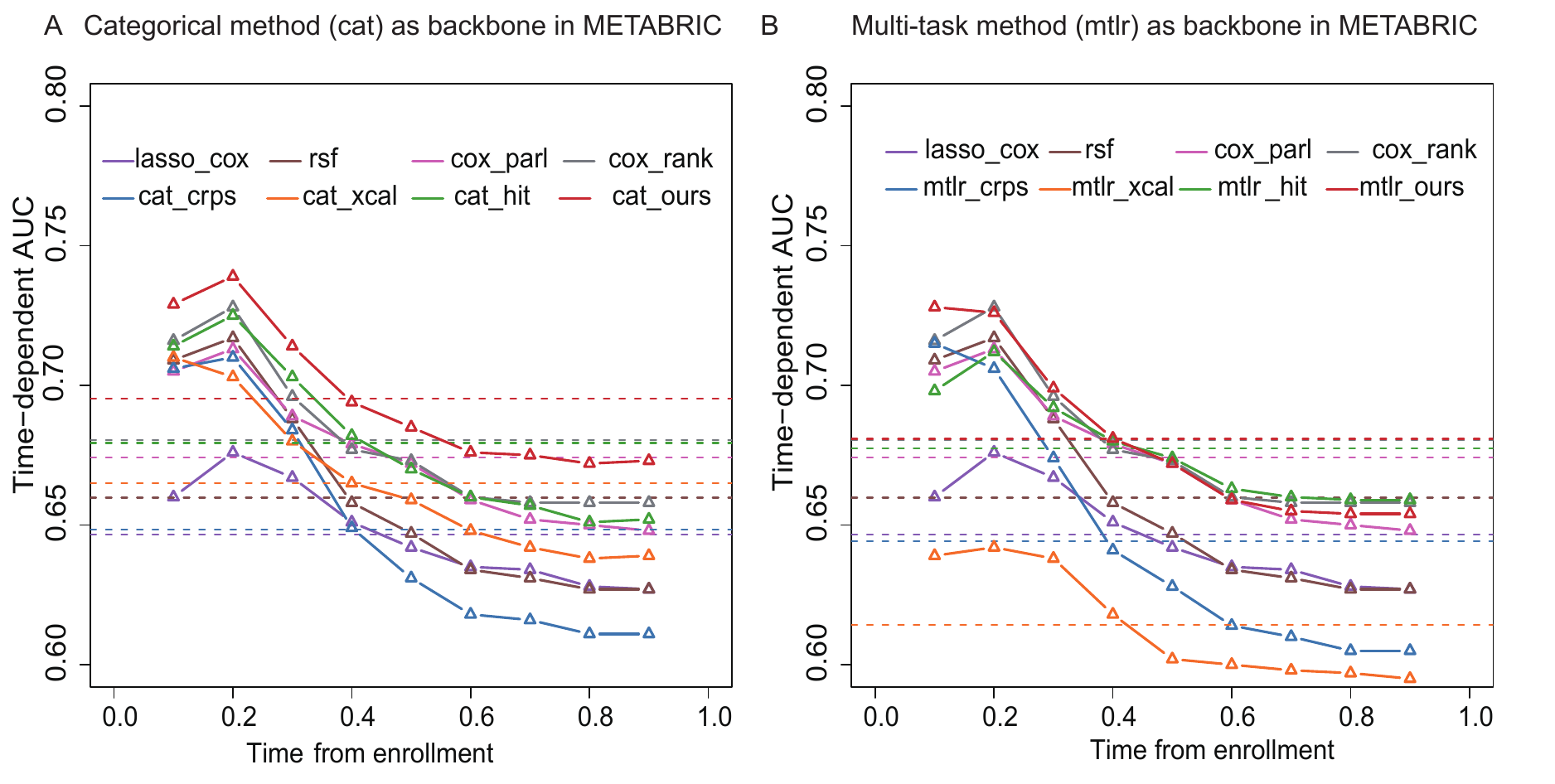}
    \caption{Performance comparison using the TDAUC in METABRIC.}
    \label{fig7-mebriAUC}
    \end{figure}

As is shown in Table \ref{tab:chap:tableMetab}, compared to other competing models, the Cat-ours achieved the highest C-index, mTDAUC (evaluation of TDAUC at each time point is shown in Figure \ref{fig7-mebriAUC}), and the lowest IBS. The same results were observed in the MTLR-ours. For survival models based on the X-cal loss, the poor performance showed a large difference between the Cat and MTLR method. In terms of risk stratification ability, the Cat-ours model identified 63\% of patients as high-risk patients with the highest risk ratio (HR=2.70) between high and low-risk patients, while the MTLR-ours model identified 53\% of patients as high-risk patients with the second highest risk ratio (HR=2.0) between high and low-risk patients. Other survival models have lower ability in identifying high-risk patients. 

\begin{table}[htbp]
	\centering
	\resizebox{\linewidth}{!}{
	\begin{tabular}[c]{ccccc}
		\toprule
		{Method} & {C-index($\uparrow$)} & {mTDAUC($\uparrow$)} & {IBS($\downarrow$)} & {HR($\uparrow$)} \\
		\midrule
		Random (ref)&	0.500& 	0.500& 	0.2500& 	1.00   \\
		Lasso-cox&	0.654&	0.647&	0.1862&	2.18 \\
		RSF&	0.674&	0.660&	0.1891&	2.46 \\
		DeepSurv&	0.670&	0.674&	0.1886&	2.36 \\
		DeepRank&	0.675&	0.680&	0.1918&	2.44 \\
		Cat-crps&	0.659&	0.648&	0.1911&	2.26 \\
		Cat-xcal& 0.660&	0.665&	0.1907&	2.27  \\
		Cat-hit&	0.674&	0.679&	0.1954&	2.17 \\
		Cat-ours&	\textbf{0.688}&	\textbf{0.695}&	0.1878&	\textbf{2.70} \\
		MTLR-crps&	0.662&	0.644&	0.1899&	2.15 \\
		MTLR-xcal&	0.612&	0.614&	0.1980&	1.86 \\
		MTLR-hit&	0.671&	0.677&	0.1874&	2.32 \\
		MTLR-ours&	0.679&	0.681&	\textbf{0.1870}&	2.50 \\
		\bottomrule
	\end{tabular}}
	\caption {Performance comparison in METABRIC}
	\label{tab:chap:tableMetab}
    \end{table}
    
\subsubsection{Experiment 4: Survival prediction in synthetic MNIST}
In this experiment, The failure time of survival MNIST is synthetic but real clinical data, so we do not perform the risk stratification.

\begin{figure}[htbp]
    \centering
    \includegraphics[width=0.47\textwidth]{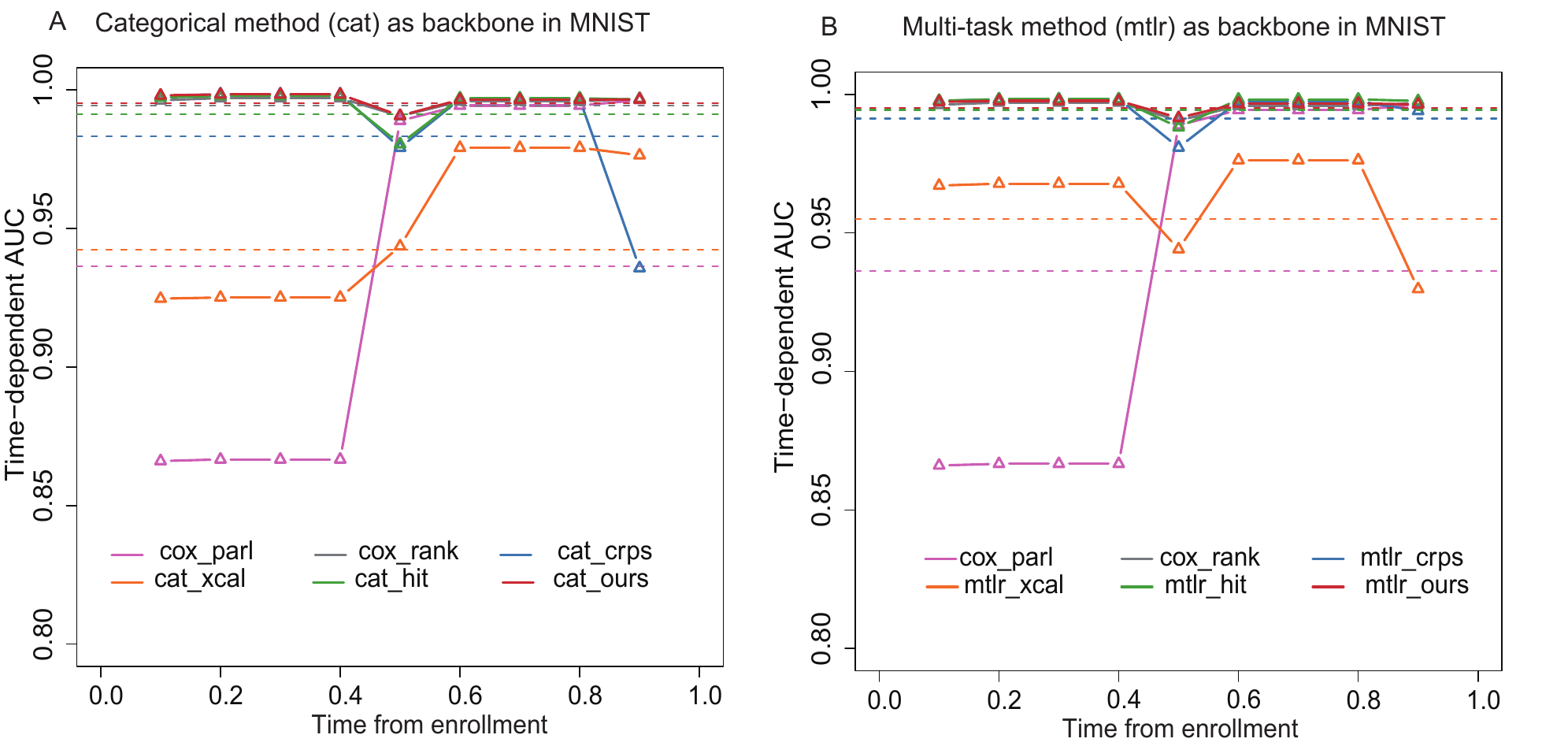}
    \caption{Performance comparison using the TDAUC in MNIST.}
    \label{fig9mnistAUC}
    \end{figure}

\begin{table}[ht]
	\centering
	\begin{tabular}[c]{ccccc}
		\toprule
		{Method} & {C-index($\uparrow$)} & {mTDAUC($\uparrow$)} & {IBS($\downarrow$)} \\
		\midrule
		Random (ref)&	0.500& 	0.500& 	0.2500\\
		DeepSurv&	0.929&	0.936&	0.0658 \\
		DeepRank&	0.951&	0.994&	0.0271 \\
		Cat-crps&	0.942&	0.983&	0.0255 \\
		Cat-xcal&	0.879&	0.942&	0.0591 \\
		Cat-hit&	0.945&	0.991&	0.0061 \\
		Cat-ours&	\textbf{0.956}&	\textbf{0.995}&	0.0060  \\
		MTLR-crps&	0.948&	0.991&	0.0226 \\
		MTLR-xcal&	0.908&	0.955&	0.0480 \\
		MTLR-hit&	0.951&	0.994&	\textbf{0.0051} \\
		MTLR-ours&	\textbf{0.956}&	\textbf{0.995}&	0.0061 \\
		\bottomrule
	\end{tabular}
	\caption {Performance comparison in MNIST}
	\label{tab:chap:Mnist}
\end{table}
The experimental results were summarized in Table \ref{tab:chap:Mnist}. Our model, using either the Category method or the MTLR method, has the highest C-index and mTDAUC compared to other models (as shown in Figure \ref{fig9mnistAUC}). The MTLR-hit have the better IBS than ours since our models take model robustness into consideration. Observing Figure \ref{fig9mnistAUC} shows that survival models based on partial likelihood and xcal combination loss have significant shortcomings in predicting failure risks at early times.

\subsection{4.4. Ablation Study}
 We conduct our ablation experiments using BIDDING dataset since it has the largest sample size among the three real-world datasets. The experiment settings are the same as those set in mentioned experiments. 
The experimental results indicate that our methods exhibit superior performance (Table \ref{tab:chap:table_7abl-cat} and \ref{tab:chap:table_8abl-mtlr}). Additionally, the model based on our TAPR loss also outperformed the models based on the existing Rank loss in these three metrics. We found that the our model(Cat-ours) had a slightly higher C-index (0.782 vs. 0.779) and similar mTDAUC (0.800 vs. 0.801) compared to the model using likelihood and TAPR-loss, but a significantly higher IBS (0.1922 vs. 0.1993). The same results were also found for the MTLR-ours. These results suggest that adding a calibration loss function to the combination of likelihood and TAPR losses can slightly improve the model's discrimination ability and its robustness. Furthermore, although the model only based on MLE showed the highest IBS, it may cause the survival model to prioritize calibration ability at the cost of discrimination ability, while our models can achieve a good balance between discrimination and calibration abilities.

\begin{table}[ht]
	\centering
	\resizebox{\linewidth}{!}{
	\begin{tabular}[c]{ccccccc}
		\toprule
		\multicolumn{4}{c}{losses}& \multicolumn{3}{c}{metrics} \\
		\midrule
		MLE	&Rank&	TAPR(ours)&	Calibration	&	C-index$(\uparrow)$&IBS$(\downarrow)$ & mTDAUC$(\uparrow)$    \\

		$\checkmark$ &	& & & 0.755&	\textbf{0.1912}&	0.778     \\
		 &$\checkmark$	& & & 0.757&0.4239&	0.770      \\
		 &	&$\checkmark$ & &0.766&	0.3038&	0.782      \\
		$\checkmark$ &$\checkmark$	& & & 0.761&	0.2061&	0.783      \\
		$\checkmark$ &	& $\checkmark$& & 0.779&	0.1993&	\textbf{0.801}      \\
		$\checkmark$ &	&$\checkmark$ &$\checkmark$ & \textbf{0.782}&	0.1922&	0.800      \\
		\bottomrule
	\end{tabular}}
	\caption {Ablation study results using the Category method}
	\label{tab:chap:table_7abl-cat}
\end{table}

\begin{table}[!ht]
	\centering
	\resizebox{\linewidth}{!}{
	\begin{tabular}[c]{ccccccc}
		\toprule
		\multicolumn{4}{c}{losses}& \multicolumn{3}{c}{metrics} \\
		\midrule
		MLE	&Rank&	TAPR(ours)&	Calibration	&	C-index$(\uparrow)$&IBS$(\downarrow)$ & mTDAUC$(\uparrow)$    \\

		$\checkmark$ &	& & & 0.763&	\textbf{0.1866}&	0.783      \\
		 &$\checkmark$	& & & 0.749	& 0.4804	&0.766       \\
		 &	&$\checkmark$ & &0.766&	0.3103&	0.786       \\
		$\checkmark$ &$\checkmark$	& & & 0.773&	0.2078&	0.791       \\
		$\checkmark$ &	& $\checkmark$& & 0.782&	0.1988&	0.797       \\
		$\checkmark$ &	&$\checkmark$ &$\checkmark$ & \textbf{0.785}&	0.1935&	\textbf{0.801}       \\
		\bottomrule
	\end{tabular}}
	\caption {Ablation study results using the MTLR method}
	\label{tab:chap:table_8abl-mtlr}
\end{table}

\section{6. Conclusion and future work}
In this study, we propose a simple yet efficient loss function, namely TripleSurv, to further optimize the modeling process of survival analysis from multiple aspects. The method is evaluated by geometrical and clinical metrics.The TripleSurv strikes a balance to facilitate the model to take into account the data distribution, ranking, and calibration. Our TripleSurv is evaluated on three real-world tasks and a semi-synthetic task. The results experimentally demonstrate that our method can enhance the discrimination and robustness of survival models against baselines including state-of-the-art models. Still,further work should be done for improvement particularly for clinical censored data modelling. For the future work, it is natural to investigate our method for survival analysis based on multi-mode data (e.g., videos, images, and diagnostic reports).

\bigskip

\bibliography{aaai23}

\begin{thebibliography}{25}
\providecommand{\natexlab}[1]{#1}

\bibitem[{Avati et~al.(2020)Avati, Duan, Zhou, Jung, Shah, and Ng}]{RN993}
Avati, A.; Duan, T.; Zhou, S.; Jung, K.; Shah, N.~H.; and Ng, A.~Y. 2020.
\newblock Countdown Regression: Sharp and Calibrated Survival Predictions.

\bibitem[{Avati et~al.(2018)Avati, Jung, Harman, Downing, Ng, and Shah}]{RN6}
Avati, A.; Jung, K.; Harman, S.; Downing, L.; Ng, A.; and Shah, N.~H. 2018.
\newblock Improving palliative care with deep learning.
\newblock \emph{BMC medical informatics and decision making}, 18(4): 55--64.

\bibitem[{Bello et~al.(2019)Bello, Dawes, Duan, Biffi, de~Marvao, Howard, Gibbs, Wilkins, Cook, and Rueckert}]{RN305}
Bello, G.~A.; Dawes, T.~J.; Duan, J.; Biffi, C.; de~Marvao, A.; Howard, L.~S.; Gibbs, J. S.~R.; Wilkins, M.~R.; Cook, S.~A.; and Rueckert, D. 2019.
\newblock Deep-learning cardiac motion analysis for human survival prediction.
\newblock \emph{Nature machine intelligence}, 1(2): 95.

\bibitem[{Cox(1972)}]{RN644}
Cox, D.~R. 1972.
\newblock Regression models and life‐tables.
\newblock \emph{Journal of the Royal Statistical Society: Series B (Methodological)}, 34(2): 187--202.

\bibitem[{Cox(1975)}]{RN9}
Cox, D.~R. 1975.
\newblock Partial likelihood.
\newblock \emph{Biometrika}, 62(2): 269--276.

\bibitem[{Gneiting and Katzfuss(2014)}]{RN5}
Gneiting, T.; and Katzfuss, M. 2014.
\newblock Probabilistic forecasting.
\newblock \emph{Annual Review of Statistics and Its Application}, 1: 125--151.

\bibitem[{Goldstein et~al.(2020)Goldstein, Han, Puli, Perotte, and Ranganath}]{RN902}
Goldstein, M.; Han, X.; Puli, A.; Perotte, A.; and Ranganath, R. 2020.
\newblock X-CAL: Explicit calibration for survival analysis.
\newblock \emph{Advances in neural information processing systems}, 33: 18296--18307.

\bibitem[{Graf et~al.(1999)Graf, Schmoor, Sauerbrei, and Schumacher}]{RN1038}
Graf, E.; Schmoor, C.; Sauerbrei, W.; and Schumacher, M. 1999.
\newblock Assessment and comparison of prognostic classification schemes for survival data.
\newblock \emph{Statistics in medicine}, 18(17‐18): 2529--2545.

\bibitem[{Harrell et~al.(1982)Harrell, Califf, Pryor, Lee, and Rosati}]{RN408}
Harrell, F.~E.; Califf, R.~M.; Pryor, D.~B.; Lee, K.~L.; and Rosati, R.~A. 1982.
\newblock Evaluating the yield of medical tests.
\newblock \emph{Jama}, 247(18): 2543--2546.

\bibitem[{He et~al.(2016)He, Zhang, Ren, and Sun}]{RNhe2016deep}
He, K.; Zhang, X.; Ren, S.; and Sun, J. 2016.
\newblock Deep residual learning for image recognition.
\newblock In \emph{Proceedings of the IEEE conference on computer vision and pattern recognition}, 770--778.

\bibitem[{Jing et~al.(2019)Jing, Zhang, Wang, Jin, Liu, Qiu, Ke, Sun, He, Hou, Tang, Lv, and Li}]{RN991}
Jing, B.; Zhang, T.; Wang, Z.; Jin, Y.; Liu, K.; Qiu, W.; Ke, L.; Sun, Y.; He, C.; Hou, D.; Tang, L.; Lv, X.; and Li, C. 2019.
\newblock A deep survival analysis method based on ranking.
\newblock \emph{Artificial Intelligence in Medicine}, 98: 1--9.

\bibitem[{Kamarudin, Cox, and Kolamunnage-Dona(2017)}]{RN1036}
Kamarudin, A.~N.; Cox, T.; and Kolamunnage-Dona, R. 2017.
\newblock Time-dependent ROC curve analysis in medical research: current methods and applications.
\newblock \emph{BMC medical research methodology}, 17(1): 1--19.

\bibitem[{Kamran and Wiens(2021)}]{RN909}
Kamran, F.; and Wiens, J. 2021.
\newblock Estimating Calibrated Individualized Survival Curves with Deep Learning.
\newblock In \emph{Proceedings of the AAAI Conference on Artificial Intelligence}, volume~35, 240--248.
\newblock ISBN 2374-3468.

\bibitem[{Katzman et~al.(2018)Katzman, Shaham, Cloninger, Bates, Jiang, and Kluger}]{RN645}
Katzman, J.~L.; Shaham, U.; Cloninger, A.; Bates, J.; Jiang, T.; and Kluger, Y. 2018.
\newblock DeepSurv: personalized treatment recommender system using a Cox proportional hazards deep neural network.
\newblock \emph{BMC medical research methodology}, 18(1): 24.

\bibitem[{Lee et~al.(2018)Lee, Zame, Yoon, and van~der Schaar}]{RN632deephit}
Lee, C.; Zame, W.~R.; Yoon, J.; and van~der Schaar, M. 2018.
\newblock Deephit: A deep learning approach to survival analysis with competing risks.
\newblock In \emph{Thirty-Second AAAI Conference on Artificial Intelligence}, volume~32.
\newblock ISBN 2374-3468.

\bibitem[{Li et~al.(2016)Li, Wang, Ye, and Reddy}]{RN1022}
Li, Y.; Wang, J.; Ye, J.; and Reddy, C.~K. 2016.
\newblock A Multi-Task Learning Formulation for Survival Analysis.
\newblock In \emph{Proceedings of the 22nd ACM SIGKDD international conference on knowledge discovery and data mining}, 1715--1724. ACM.

\bibitem[{Rajkomar et~al.(2018)Rajkomar, Oren, Chen, Dai, Hajaj, Hardt, Liu, Liu, Marcus, and Sun}]{RN7}
Rajkomar, A.; Oren, E.; Chen, K.; Dai, A.~M.; Hajaj, N.; Hardt, M.; Liu, P.~J.; Liu, X.; Marcus, J.; and Sun, M. 2018.
\newblock Scalable and accurate deep learning with electronic health records.
\newblock \emph{NPJ digital medicine}, 1(1): 1--10.

\bibitem[{Ranganath et~al.(2016)Ranganath, Perotte, Elhadad, and Blei}]{RN1021}
Ranganath, R.; Perotte, A.; Elhadad, N.; and Blei, D. 2016.
\newblock Deep survival analysis.
\newblock In \emph{Machine Learning for Healthcare Conference}, 101--114. PMLR.

\bibitem[{Raykar et~al.(2007)Raykar, Steck, Krishnapuram, Dehing-Oberije, and Lambin}]{RN360}
Raykar, V.~C.; Steck, H.; Krishnapuram, B.; Dehing-Oberije, C.; and Lambin, P. 2007.
\newblock On Ranking in Survival Analysis: Bounds on the Concordance Index.
\newblock In \emph{Conference on Advances in Neural Information Processing Systems}.

\bibitem[{Ren et~al.(2019)Ren, Qin, Zheng, Yang, Zhang, Qiu, and Yu}]{RN651}
Ren, K.; Qin, J.; Zheng, L.; Yang, Z.; Zhang, W.; Qiu, L.; and Yu, Y. 2019.
\newblock Deep recurrent survival analysis.
\newblock In \emph{Proceedings of the AAAI Conference on Artificial Intelligence}, volume~33, 4798--4805.
\newblock ISBN 2374-3468.

\bibitem[{Shah, Steyerberg, and Kent(2018)}]{RN16}
Shah, N.~D.; Steyerberg, E.~W.; and Kent, D.~M. 2018.
\newblock Big data and predictive analytics: recalibrating expectations.
\newblock \emph{Jama}, 320(1): 27--28.

\bibitem[{Steck et~al.(2007)Steck, Krishnapuram, Dehing-Oberije, Lambin, and Raykar}]{RN8}
Steck, H.; Krishnapuram, B.; Dehing-Oberije, C.; Lambin, P.; and Raykar, V.~C. 2007.
\newblock On ranking in survival analysis: Bounds on the concordance index.
\newblock \emph{Advances in neural information processing systems}, 20.

\bibitem[{Tibshirani(1997)}]{RN10}
Tibshirani, R. 1997.
\newblock The lasso method for variable selection in the Cox model.
\newblock \emph{Statistics in medicine}, 16(4): 385--395.

\bibitem[{Van~Calster and Vickers(2015)}]{RN17}
Van~Calster, B.; and Vickers, A.~J. 2015.
\newblock Calibration of risk prediction models: impact on decision-analytic performance.
\newblock \emph{Medical decision making}, 35(2): 162--169.

\bibitem[{Wang, Li, and Chignell(2021)}]{RN12}
Wang, L.; Li, Y.; and Chignell, M. 2021.
\newblock Combining Ranking and Point-wise Losses for Training Deep Survival Analysis Models.
\newblock In \emph{2021 IEEE International Conference on Data Mining (ICDM)}, 689--698. IEEE.
\newblock ISBN 1665423986.

\end{thebibliography}

\end{document}